\documentclass{article}

\usepackage{arxiv}

\usepackage[utf8]{inputenc} % allow utf-8 input
\usepackage[T1]{fontenc}    % use 8-bit T1 fonts
\usepackage{hyperref}       % hyperlinks
\usepackage{url}            % simple URL typesetting
\usepackage{booktabs}       % professional-quality tables
\usepackage{amsfonts}       % blackboard math symbols
\usepackage{nicefrac}       % compact symbols for 1/2, etc.
\usepackage{microtype}      % microtypography
\usepackage{lipsum}
\usepackage{graphicx}
\graphicspath{ {./images/} }
\usepackage{amsmath}

\title{PathSegDiff: Pathology Segmentation using Diffusion model representations}

\author{
 Sachin Kumar Danisetty \\
  Stony Brook University\\
  Stony Brook, NY \\
  \texttt{sdanisetty@cs.stonybrook.edu} \\
   \And
 Alexandros Graikos \\
  Stony Brook University\\
  Stony Brook, NY \\
  \texttt{agraikos@cs.stonybrook.edu} \\
  \And
 Srikar Yellapragada \\
  Stony Brook University\\
  Stony Brook, NY \\
  \texttt{myellapragad@cs.stonybrook.edu} \\
  \And
 Dimitris Samaras \\
  Stony Brook University\\
  Stony Brook, NY \\
  \texttt{samaras@cs.stonybrook.edu} \\
}

\begin{document}
\maketitle
\begin{abstract}
Image segmentation is crucial in many computational pathology pipelines, including accurate disease diagnosis, subtyping, outcome, and survivability prediction. The common approach for training a segmentation model relies on a pre-trained feature extractor and a dataset of paired image and mask annotations. These are used to train a lightweight prediction model that translates features into per-pixel classes. The choice of the feature extractor is central to the performance of the final segmentation model, and recent literature has focused on finding tasks to pre-train the feature extractor. In this paper, we propose PathSegDiff, a novel approach for histopathology image segmentation that leverages Latent Diffusion Models (LDMs) as pre-trained featured extractors. Our method utilizes a pathology-specific LDM, guided by a self-supervised encoder, to extract rich semantic information from H\&E stained histopathology images. We employ a simple, fully convolutional network to process the features extracted from the LDM and generate segmentation masks. Our experiments demonstrate significant improvements over traditional methods on the BCSS and GlaS datasets, highlighting the effectiveness of domain-specific diffusion pre-training in capturing intricate tissue structures and enhancing segmentation accuracy in histopathology images.
\end{abstract}

% keywords can be removed

\section{Introduction}
Histopathology involves the microscopic study of tissue structures to diagnose diseases, particularly cancer. Whole Slide Images (WSIs), characterized by their gigapixel resolution, serve as a rich data source, offering detailed visualizations of tissue morphology. However, tasks like semantic segmentation, which involve identifying tumor regions and tissue subtypes, remain challenging due to the need for dense, expert annotations. This challenge is further compounded by the heterogeneity of histopathological features, such as varying tissue morphologies. To alleviate the reliance on exhaustive manual labeling, there is a growing shift towards adapting pre-trained supervised or unsupervised models. These models learn robust and generalizable representations and, with limited expert annotations, can perform the downstream segmentation task. 

Recent advances in generative modeling, particularly diffusion models \cite{ho2020denoising}, have shown that the representations of the denoiser network are highly efficient in pixel-level prediction tasks \cite{baranchuk}, making them an appealing choice to use as the feature extraction model. The ability to generate coherent images from noise demonstrates a strong grasp of image structure and spatial relationships, making them suited for tasks, like segmentation, that require rich semantic understanding. With diffusion models now having also been adapted to the histopathology domain \cite{yellapragada2024pathldm}, in this paper we explore their application to histopathology segmentation tasks.

In this work, we introduce \textbf{PathSegDiff}, a novel approach that addresses the challenges of histopathology segmentation by leveraging a Latent Diffusion Model (LDM) pre-trained on histopathology datasets as a feature extractor. Unlike methods that rely on supervised pre-training, PathSegDiff utilizes the semantically rich features learned by a diffusion model to perform the segmentation. Following feature extraction, our method employs a fully convolutional network (FCN) for mask generation, enabling precise identification of tumor regions and other critical tissue types. PathSegDiff achieves state-of-the-art performance on gland segmentation, significantly surpassing ResNet-18 baselines in quantitative and qualitative evaluation metrics. These results highlight the importance of utilizing domain-specific knowledge from generative models to advance segmentation tasks in computational pathology.

\section{Related Work}
\subsection{Histopathology Image Segmentation}
In recent years, deep learning approaches have demonstrated remarkable success in histopathology image segmentation, particularly for H\&E stained samples. The field has seen many methods ranging from fully supervised to weakly supervised and unsupervised techniques, as comprehensively reviewed by Srinidhi et al. \cite{srinidhi}. Among these, U-Net-based architectures have emerged as particularly effective, leveraging skip connections to mitigate vanishing gradients and enable richer feature extraction \cite{unet}. Weakly-supervised learning methods have gained traction due to the challenges of obtaining pixel-level annotations in histopathology. For instance, Xu et al. proposed multiple clustered instance learning (MCIL), which simultaneously performs image-level classification, segmentation, and patch-level clustering \cite{mcil}. Similarly, Liu et al. introduced a framework based on sparse patch annotation for tumor segmentation \cite{sparse}. Fully convolutional networks (FCNs) have also shown promise in histopathological gland segmentation. Chen et al. developed a deep contour-aware network that effectively addresses key challenges in gland delineation \cite{dcan}. These approaches often incorporate multi-scale feature extraction and boundary refinement techniques to improve segmentation accuracy.

The issue of class imbalance, particularly prevalent in histopathology datasets, has been addressed through various sampling and augmentation strategies. Bokhorst et al. compared instance-based and mini-batch-based balancing approaches when working with sparse annotations \cite{bokhorst}, highlighting the importance of data handling techniques in improving model performance. Recent advancements in weakly supervised segmentation for whole slide images (WSIs) have focused on CAM-based approaches. Methods like Histo-Seg \cite{histoseg} utilize gradient-weighted CAM with post-processing, while SEAM \cite{seam} enforces semantic consistency across views to improve feature space compactness. Other approaches, such as C-CAM \cite{CCAM} with causal chains, WSSS-Tissue \cite{wsss} with progressive pseudo-supervision, and PistoSeg \cite{pistoseg} with Mosaic transformations, demonstrate innovative strategies for generating pseudo-masks and improving segmentation accuracy.

Furthermore, Yan et al. \cite{yan} developed a multi-scale encoder network to enhance feature extraction specific to pathology. In contrast, Yang et al. \cite{yang} presented a deep metric learning-based retrieval method incorporating mixed attention mechanisms for improved semantic similarity metrics. Chen et al. leverage hierarchical structures for multi-scale vision transformers \cite{vit} using self-supervised learning to learn high-resolution image representations \cite{hipt}. Zhang et al. enhance the Segment Anything Model's (SAM) ability to conduct semantic segmentation in pathology by incorporating the encoder of the HIPT model as an external pathology encoder \cite{sam-path}. 

\subsection{Histopathology Diffusion Models}
Diffusion models have been solidified in digital histopathology through works that showed their ability to model the underlying image distributions \cite{yellapragada2024pathldm}. As diffusion models are known to require conditioning signals to guide the generation process, prior works in pathology have leveraged self-supervised learning (SSL) embeddings as effective conditioning signals for diffusion models \cite{graikos2024learned}. While these models have successfully captured domain-specific features, their applications have primarily been limited to data augmentation and classification, and the use of these models for segmentation remains underexplored.

\subsection{Diffusion-based Image Segmentation}
Diffusion probabilistic models have emerged as a powerful tool in the realm of image segmentation, particularly within the medical imaging domain. These models operate through a forward process where input data is progressively perturbed with noise, followed by a reverse process where the noise is systematically removed to reconstruct the original data. This approach has been leveraged for various tasks including image generation, translation, reconstruction, classification, and segmentation, showcasing their versatility and effectiveness in handling complex data distributions.

Amit et al. \cite{amit2021segdiff} pioneered the application of diffusion models to image segmentation by using a conditional diffusion model. Their work laid the groundwork for subsequent research in this area. Building on this, Baranchuk et al. \cite{baranchuk} demonstrated that diffusion models can capture high-level semantic information, which is crucial for label-efficient semantic segmentation. They utilized ensembles of multi-layer perceptions (MLPs) for pixel-wise classification, enhancing the segmentation process.

Wolleb et al. \cite{wolleb} further advanced the field by employing a stochastic sampling process within diffusion models to generate a distribution of segmentation masks, thereby addressing the inherent ambiguity in medical image segmentation. Kim et al. \cite{kim} introduced a novel approach combining diffusion models with adversarial learning for vessel segmentation, highlighting the potential of integrating different learning paradigms to improve segmentation accuracy. CIMD \cite{cimd}, introduced in, captures the heterogeneity of segmentation masks without requiring an additional network for prior information during inference.

Wu et al. \cite{medsegdiff} developed MedSegDiff, a diffusion-based model tailored for general medical image segmentation. They introduced a dynamic conditional encoding strategy to manage step-wise attention and proposed techniques to eliminate high-frequency noise components, thereby improving segmentation quality. An evolution of this model, MedSegDiffV2 \cite{medsegdiff2}, incorporated Transformer-based architectures, enhancing the conditioning techniques over the backbone using raw images in the diffusion process.

Kazerouni et al. \cite{kazerouni} offered an extensive overview of diffusion models in medical imaging, emphasizing their application across different modalities like CT, MRI, and ultrasound. Our framework distinguishes itself by focusing on the unique challenges of medical image segmentation, particularly in handling the inherent uncertainty and variability in medical data, which often requires an ensemble of predictions to achieve robust results.

\section{Method}
\subsection{Pre-trained Diffusion Model}
The first part of our approach is pre-training the diffusion generative model on histopathology data. Instead of training the model from scratch on the datasets we are interested in performing segmentation on, which are limited in the number of images they provide, we utilize the model of Graikos et al. \cite{graikos2024learned} which is trained on the entirety of TCGA-BRCA \cite{tcga}. The model we use is a latent diffusion model \cite{ldm} which consists of three primary components: a Variational Autoencoder (VAE) for image compression into latent representations, a UNet-based denoiser that learns the diffusion process from Gaussian noise to latents, and a cross-attention conditioning mechanism. The diffusion model relies on a separate, self-supervised encoder (in our case HIPT \cite{hipt}) for conditioning. Conditioning the model on self-supervised embeddings is shown to improve its performance by providing the necessary global structural coherence and spatial consistency context \cite{graikos2024learned}.

The SSL-guided LDM's UNet architecture incorporates convolution blocks, multi-scale sampling operations, residual connections, and attention modules that facilitate cross-attention between SSL embeddings and UNet features. During denoising, the model leverages input information to determine optimal denoising trajectories, while cross-attention layers establish semantic correlations between visual features and input data, yielding rich discriminative representations. We aim to exploit the internal workings of the UNet and extract useful features from its decoder module that we will then use to train a segmentation head network.

\subsection{Feature Extraction}
Although diffusion models utilize an iterative process to synthesize an image, our methodology requires only a single forward pass through the diffusion model for visual representation extraction, circumventing the complete generative process. 

Formally, given an input image $x \in \mathbb{R}^{H \times W \times 3}$, we first obtain the conditioning vector through the self-supervised feature extractor $SSL$:
\begin{equation}
    y = SSL(x)
\end{equation}
To extract features from the diffusion UNet we will provide this conditioning along with the noisy image that the network expects to denoise.
Since we are using a Latent Diffusion Model \cite{rombach2022high}, the input image is first processed by the LDM's encoder $\mathcal{E}$ to generate its latent representation $z_0$. The encoder downsamples the image by a factor of 8, embedding the image patches into 4-dimensional vectors
\begin{equation}
    z_0 = \mathcal{E}(x), \quad z_0 \in \mathbb{R}^{H/8 \times W/8 \times 4}
\end{equation}
The noisy latent $z_t$ is then obtained at timestep $t$ through the forward diffusion process:
\begin{equation}\label{noise_latent}
    z_t = \sqrt{\bar{\alpha_t}}z_0 + \sqrt{1 - \bar{\alpha_t}}\epsilon,\quad \epsilon \sim \mathcal{N}(0,I)
\end{equation}
Here, $t$ represents the diffusion step, and $\alpha_1,...,\alpha_T$ define the diffusion noise schedule where $\bar{\alpha_t} = \prod_{k=1}^{t}\alpha_k$, as established in DDPM \cite{ho2020denoising}. We then give the conditioning vector $y$ and the noisy latent $z_t$ to the diffusion UNet and extract intermediate features. Specifically, we extract the activations from the middle block and all upsampling blocks of the denoiser UNet. Since these UNet representations are of different spatial dimensions, we upsample them using bilinear interpolation to $H \times W$ dimensions, enabling their interpretation as pixel-level representations of $x$:
\begin{equation}
    f = \text{BilinearInterp}(\text{UNet}_{\text{middle}}(z_t,y), \text{UNet}_{\text{upsample}_1}, \dots, \text{UNet}_{\text{upsample}_{12}}(z_t,y))
\end{equation}

\begin{figure}
    \centering
    \includegraphics[width=.7\linewidth]{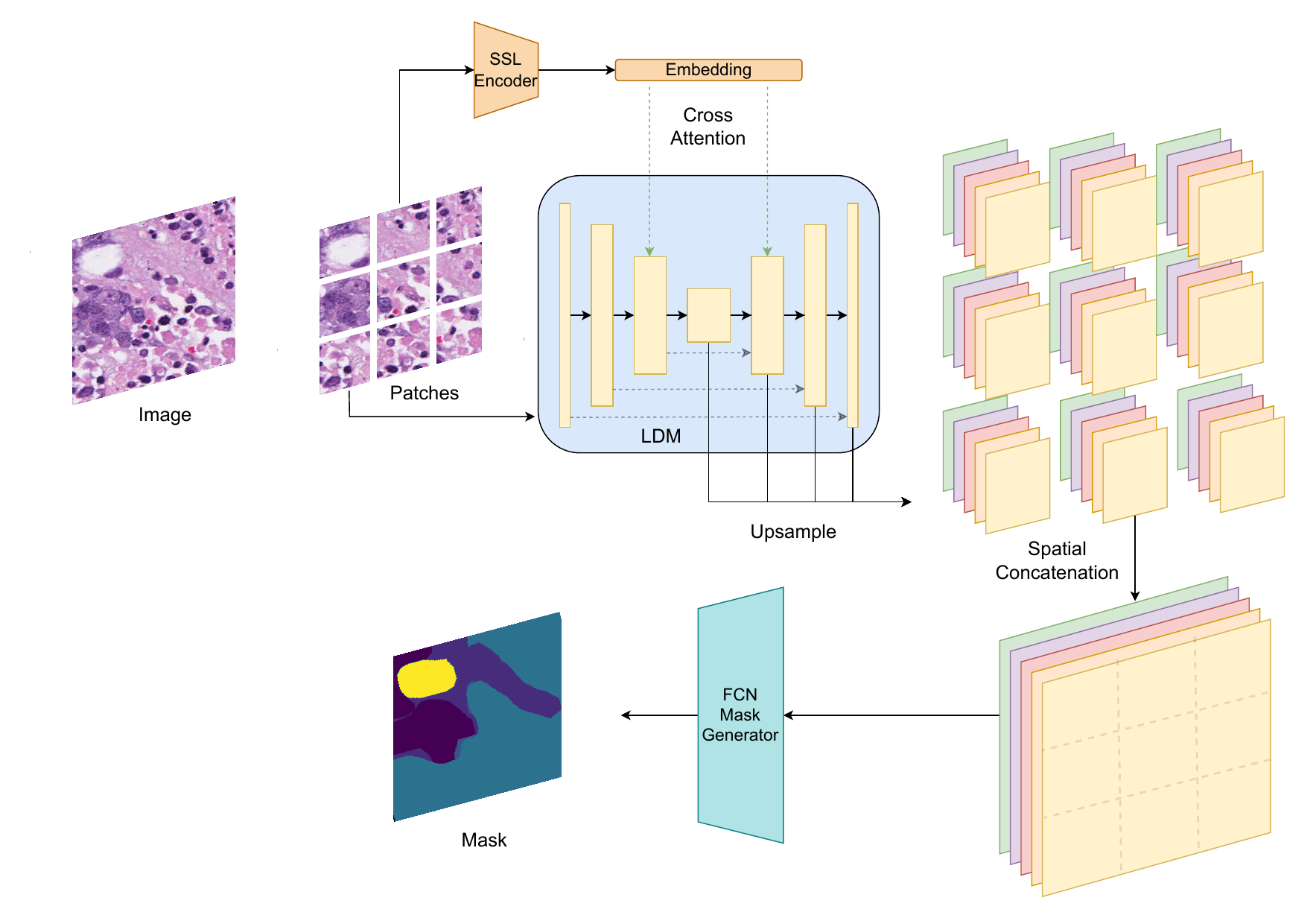}
    \caption{\textbf{Overview of PathSegDiff:} (1) We partition an image into non-overlapping patches, (2) Using the SSL encoder-based conditioning we extract features from the LDM U-Net decoder, aligning spatially the features by upsampling (3) We spatially concatenate the features to create a full representation of the image (4) We use an FCN network to predict the segmentation maps.}
    \label{arch}
\end{figure}

\subsection{Mask Generation}

The mask generator architecture comprises a fully convolutional network (FCN) that processes the UNet features $f$ to generate pixel-wise class probabilities. The FCN architecture incorporates a sequence of two convolution layers designed for channel dimensionality reduction, with two intermediate 2D transposed convolution layers facilitating the upsampling operations. The resulting mask maintains the same dimensions as the original input image. The mask generation process can be formally expressed as:
\begin{equation}
     m = FCN(f) , \quad m \in \mathbb{R}^{H \times W \times K}
\end{equation}
where $H$ represents the FCN mapping function and $m$ denotes the output probability masks over the K predicted classes.

During the training phase, we maintain the pre-trained weights of both the HIPT encoder and LDM frozen while optimizing only the FCN segmentation head using pixel-wise cross-entropy loss. This approach leverages the robust feature representations learned by the frozen components while adapting the segmentation-specific layers.

To address the unique challenges of histopathology image analysis, where contextual information is crucial for accurate tissue classification, we developed a novel patch-based processing strategy. While 256×256 pixel patches proved insufficient for capturing the necessary spatial semantic relationships between tumor and surrounding cells, processing larger images directly through LDMs would be computationally prohibitive. Our solution partitions 768×768 pixel images into nine 256×256 pixel patches. Each patch undergoes feature extraction through the SSL encoder and LDM independently. The resulting UNet features are then spatially concatenated channel-wise, preserving their relative positions to construct a comprehensive feature representation of the entire image. This consolidated feature map serves as input to the FCN segmentation head, which generates the final class-specific segmentation masks. An overview of our proposed method is shown in Figure~\ref{arch}.

\section{Experiments and Results}
\subsection{Data}
For our experimental evaluation, we use the BCSS \cite{amgad2019structured}, and GlaS \cite{sirinukunwattana2017gland} datasets. The BCSS dataset is a part of the Breast Cancer Segmentation Grand Challenge. This dataset comprises over 20,000 semantic segmentation annotations derived from 151 hematoxylin and eosin H\&E stained histological breast cancer images at $20\times$ magnification, sourced from TCGA-BRCA. The original dataset includes annotations for 22 distinct classes, which we consolidated into five broader region categories: tumor, stroma, inflammatory infiltration, necrosis, and other. This mapping was performed to simplify the classification task and focus on clinically relevant tissue types. Each region of interest (ROI) was divided into overlapping image patches of $800\times800$ pixels. To focus more on relevant tissue regions, patches where more than 90\% of the area was classified as "don't care" were excluded from further processing. A random crop of $768\times768$ pixels was dynamically extracted from each valid patch. We report our evaluation on the official validation set.

The GlaS dataset is a part of the Gland Segmentation in Colon Histology Images Challenge Content held at MICCAI 2015. The dataset consists of 165 images derived from 16 H\&E stained histological sections of stage T3 or T4 colorectal adenocarcinoma. The datasets contains annotated glandular boundaries, categorizing regions to malignant and benign classes. The dataset was officially split into $train$ (85 images), $test\_A$(60 images) and $test\_B$(20 images) subsets. We report our evaluation on $test\_B$ dataset. All images are resized to $768\times768$ pixels before processing. For both datasets, we use their official training and validation splits.

\subsection{Experimental Setup}
 We employ the HIPT patch-level Vision Transformer $\text{ViT}_{256}\text{-}16$ for extracting the LDM conditioning vectors, which is pre-trained on the PanCancer TCGA dataset at $20\times$ scale using a self-supervised DINO framework. We leverage the Variational Autoencoder (VAE) and UNet components from the SSL-guided Latent Diffusion Model, pre-trained on generating $256\times256$ pixel patches of TCGA-BRCA and TCGA-CRC datasets, conditioned with embeddings extracted from the HIPT model. The weights and hyperparameters for these models are adopted directly from their respective original implementations as provided by the authors.
 
 The Fully Convolutional Network (FCN) architecture used for segmentation begins consists of 4 convolutional layers . It begins with a Conv2d layer, followed by two ConvTranspose2d layers for upsampling, and concludes with a final Conv2d layer that assigns each pixel to one of the predefined target classes. Batch normalization and ReLU activation functions are applied after each layer, except for the final classification layer, to ensure stable training dynamics and effective non-linear feature extraction.

The sizes of the models employed in this architecture are detailed in Table \ref{tab:model_sizes}.

\begin{table}[ht]
\centering
\begin{tabular}{|l|c|}
\hline
\textbf{Model} & \textbf{Number of Parameters} \\ \hline
HIPT & 21,665,664 \\ \hline
SSL-guided LDM & 478,619,777 \\ \hline
FCN & 17,470,150 \\ \hline
\end{tabular}
\vspace{5pt}
\caption{Model sizes used in the architecture.}
\label{tab:model_sizes}
\end{table}

 We select $t=50$ as the diffusion timestep at which we extract features. To optimize the segmentation head we use Adam optimizer with the learning rate set at 0.0001. To address the issue of class imbalance in the BCSS dataset, we employed a weighted categorical cross-entropy loss function. The weights were assigned based on the frequency of pixels: 
$$W_c = \begin{cases} 
    0 & : \text{if "don't care" class} \\
    1 - \frac{N_c}{N} & : \text{if otherwise}
\end{cases}$$
where $N_c$ is the number of pixels belonging to class $c$ in training dataset and $N$ is the total number of pixels in training dataset(excluding "don't care" class).

In our evaluation, we compare the proposed method with various backbone feature extraction models to assess its effectiveness in histopathology image segmentation. Specifically, we utilize features from the self-supervised HIPT model, which also serves as conditioning for the diffusion UNet, alongside ImageNet pre-trained models such as VGG16 and UNet with a ResNet-18 backbone. For BCSS, VGG16-FCN8 was used as the baseline model, while for GlaS, the benchmark performance was based on its best submission model, CUMedVision2. 

To further investigate the impact of self-supervised embeddings on feature conditioning in PathSegDiff, we compare the performance of the SSL-guided LDM with an unconditional LDM pre-trained without self-supervised embeddings. This comparative analysis highlights the advantages of leveraging domain-specific self-supervised representations for enhancing segmentation accuracy and robustness.

\subsection{Results}

\subsubsection{Quantitative Analysis}

Table \ref{tab:metrics_bcss} summarizes the performance of various models on the BCSS dataset. The baseline model, VGG16-FCN8, achieves an accuracy of 0.786 and a Dice score of 0.567, reflecting its limited ability to capture complex tissue structures due to its reliance on ImageNet pre-training. Similarly, the UNet with ResNet-18 backbone, another ImageNet-pretrained architecture, performs slightly worse with a Dice score of 0.529, indicating its challenges in extracting meaningful histopathological features.

The self-supervised HIPT model, pre-trained on the TCGA dataset, achieves significantly lower performance. This is likely due to its discriminative self-supervised training, which does not explicitly encourage learning per-pixel features but rather a single patch-level representation. In contrast, both PathSegDiff models demonstrate superior performance due to their domain-specific pre-training on the TCGA-BRCA dataset. The unconditional PathSegDiff model (PathSegDiff-uncond) achieves an accuracy of 0.815 and a Dice score of 0.757, showcasing its ability to effectively model complex tissue structures. The PathSegDiff model further improves these metrics with an accuracy of 0.826 and a Dice score of 0.781, highlighting the advantage of self-supervised embeddings in providing rich semantic information for more precise segmentation.

\begin{table}[ht]
\centering
\small
\begin{tabular}{lcccc}
\hline
& Accuracy & Dice & mIoU & F1 Score\\
\hline
HIPT & 0.619 & 0.353 & 0.321 & 0.170\\
Res-UNet & 0.751 & 0.529 & 0.524 & 0.658\\
VGG16-FCN8 & 0.786 & 0.567 & 0.577 & 0.724\\
\hline
PathSegDiff-uncond & 0.815 & 0.757 & 0.616 & 0.757\\
PathSegDiff & \textbf{0.826} & \textbf{0.781} & \textbf{0.647} & \textbf{0.780}\\
\hline
\\
\end{tabular}
\caption{Quantitative Analysis on BCSS dataset}
\label{tab:metrics_bcss}
\end{table}

Table \ref{tab:sota_bcss} compares our method against recent state-of-the-art models on the top four classes (tumor, stroma, lymphocytic infiltrate, necrosis) in the BCSS dataset using mIoU as the evaluation metric\cite{wu2025superpixelboundarycorrectionweaklysupervised}. Our PathSegDiff achieves the best overall mIoU and outperforms all other methods across most categories while maintaining competitive performance in stroma segmentation (ranking third). These results demonstrate that our method achieves an excellent balance between segmentation accuracy across different classes and surpasses strong baselines such as PistoSeg, which utilizes extensive data augmentation.

\begin{table}[ht]
\centering
\begin{tabular}{lccccc}
\hline
Model & TUM & STR & LYM & NEC & mIOU \\
\hline
HistoSegNet & 0.3314 & 0.4646 & 0.2905 & 0.0191 & 0.2764 \\
SEAM & 0.7437 & 0.6216 & 0.5079 & 0.4843 & 0.5894 \\
C-CAM & 0.7557 & 0.6796 & 0.3100 & 0.4943 & 0.5599 \\
WSSS-Tissue & 0.7798 & 0.7295 & 0.6098 & 0.6687 & 0.6970 \\
PistoSeg & 0.8110 & \textbf{0.7504} & 0.6184 & 0.6422 & 0.7055 \\
\hline
PathSegDiff & \textbf{0.8220} & 0.6998 & \textbf{0.6205} & \textbf{0.6976} & \textbf{0.7100} \\
\hline
\\
\end{tabular}
\caption{Comparison of different SOTA models on top 4 classes of BCSS dataset}
\label{tab:sota_bcss}
\end{table}

The results on the GLaS dataset are presented in Table \ref{tab:metrics_glas}. Following a similar trend to BCSS, the PathSegDiff outperforms all other models with an Accuracy of 0.90, an mIoU of 0.81, and a Dice score of 0.89, surpassing the best submission for the GLaS contest (CUMedVision2). The PathSegDiff-uncond also achieves competitive results, highlighting the effectiveness of pre-trained LDMs for segmentation tasks. In contrast, traditional models like UNet and self supervised HIPT exhibit lower performance across all metrics.

Overall, the proposed method's ability to effectively utilize domain-specific pre-training results in more precise and reliable segmentation outcomes across both datasets. This highlights the importance of tailoring pre-training strategies to suit specific medical imaging tasks.

\begin{table}[ht]
\centering
\small
\begin{tabular}{lcccc}
\hline
& Accuracy & Dice & mIoU & F1 Score \\
\hline
HIPT & 0.745 & 0.742 & 0.591 & 0.374 \\
Res-UNet & 0.778 & 0.771 & 0.630 & 0.765 \\
CUMedVision2 & - & 0.868 & -  & 0.887\\
\hline
DDPMSeg & - & - & 0.654 & 0.759 \\
CIMD & - & - & 0.569 & 0.692 \\
MedSegDiff & - & - & 0.778 & 0.852 \\
\hline
PathSegDiff-uncond & 0.896 & 0.893 & 0.807 & 0.890\\
PathSegDiff & \textbf{0.900} & \textbf{0.896} & \textbf{0.812} & \textbf{0.894}\\
\hline
\\
\end{tabular}
\caption{Quantitative Analysis on GlaS dataset}
\label{tab:metrics_glas}
\end{table}

\subsubsection{Qualitative Analysis}

\begin{figure}
    \centering
    \includegraphics[width=1\linewidth]{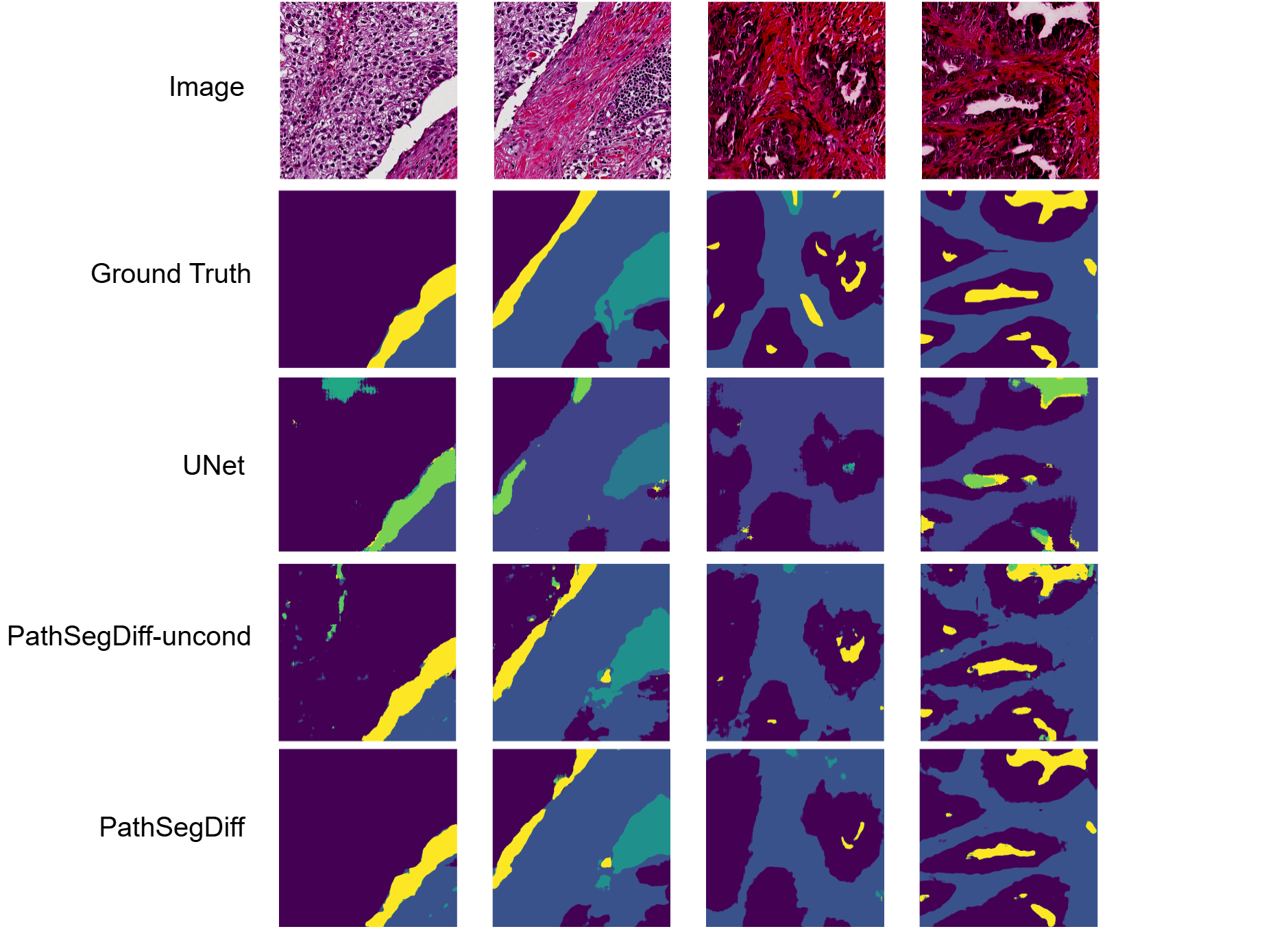}
    \caption{\textbf{Qualitative Analysis on BCSS dataset:} The top row displays original H\&E stained histopathology images, followed by ground truth segmentation masks, and predictions by UNet and our proposed methods. Our approach demonstrates better performance in accurately delineating glandular structures, closely aligning with ground truth annotations.}
    \label{bcss}
\end{figure}

\begin{figure}
    \centering
    \includegraphics[width=1\linewidth]{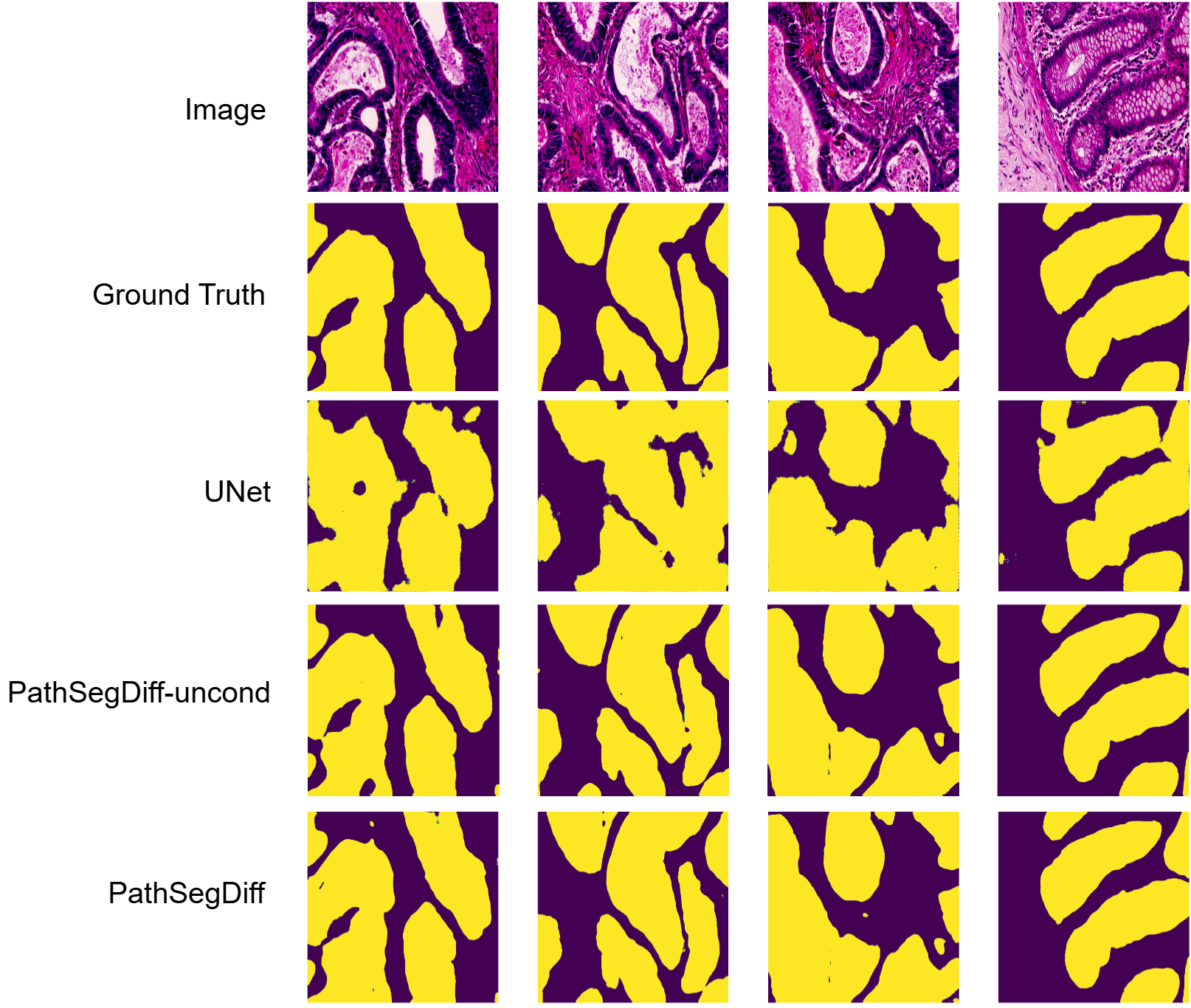}
    \caption{\textbf{Qualitative Analysis on GlaS dataset:} The top row displays original H\&E stained histopathology images, followed by ground truth segmentation masks, and predictions by UNet and our proposed methods}
    \label{glas}
\end{figure}

Our qualitative analysis demonstrates the effectiveness of the proposed PathSegDiff method across two distinct histopathology datasets. On the GlaS dataset (Fig. \ref{glas}), our methods exhibits remarkable accuracy in gland segmentation, closely adhering to the ground truth boundaries with minimal artifacts. The segmentation maps generated by our approaches maintain structural coherence and precise delineation of glandular regions, whereas the UNet, shows inconsistencies in boundary preservation and occasional over-segmentation artifacts.

In the more complex BCSS dataset (Fig. \ref{bcss}), which presents challenging multi-class tissue segmentation scenarios, our method's superiority becomes more pronounced. The proposed approaches accurately distinguishes between different tissue types while maintaining spatial consistency across region boundaries. This performance can be attributed to the domain-specific knowledge encoded in our LDM pre-training on pathology data, contrasting with ResNet-18's generic ImageNet pre-training in UNet. The traditional model struggles with fine-grained tissue differentiation, particularly in regions with subtle transitions between different tissue types, often producing fragmented or imprecise segmentation boundaries. These results underscore the importance of domain-specific pre-training in histopathology image analysis tasks.

\subsection{Ablation studies}
In this section we present the ablation study of model across various parameters.
\subsubsection{Ablation Study on Learning rate}
We conducted experiments on ablating the learning rate used for Adam optimizer to train the FCN and report the results in table \ref{tab:learning_rates}. We observe that a learning rate of $1e-4$ outperforms others. 
\begin{table}[ht]
\centering
\begin{tabular}{|c|c|c|c|}
\hline
Learning rate & Val Accuracy & Val mIOU & Val Dice \\
\hline
1e-3 & 0.819 & 0.632 & 0.769 \\
\hline
1e-4 & 0.826 & 0.646 & 0.780 \\
\hline
1e-5 & 0.821 & 0.638 & 0.774 \\
\hline
\end{tabular}
\vspace{5pt}
\caption{Validation metrics for different learning rates}
\label{tab:learning_rates}
\end{table}
\subsubsection{Diffusion Timesteps Analysis}
We also explored the effect of sampling Gaussian diffusion timesteps during feature extraction (Equation \ref{noise_latent}). Our experiments reveal that features extracted from early-middle timesteps (e.g., $t=50$) yield superior discriminative performance compared to low or high timesteps. At low timesteps, diffusion models emphasize stochastic details over structural information, while at high timesteps, feature quality degrades due to reduced recognizability of the input data. The validation metrics over different timesteps are presented in Fig. \ref{fig:timestep}
\subsubsection{Ablation Study on UNet layers}
\begin{figure}[t] 
    \centering
    \begin{minipage}{0.45\textwidth}
        \centering
        \includegraphics[width=\linewidth]{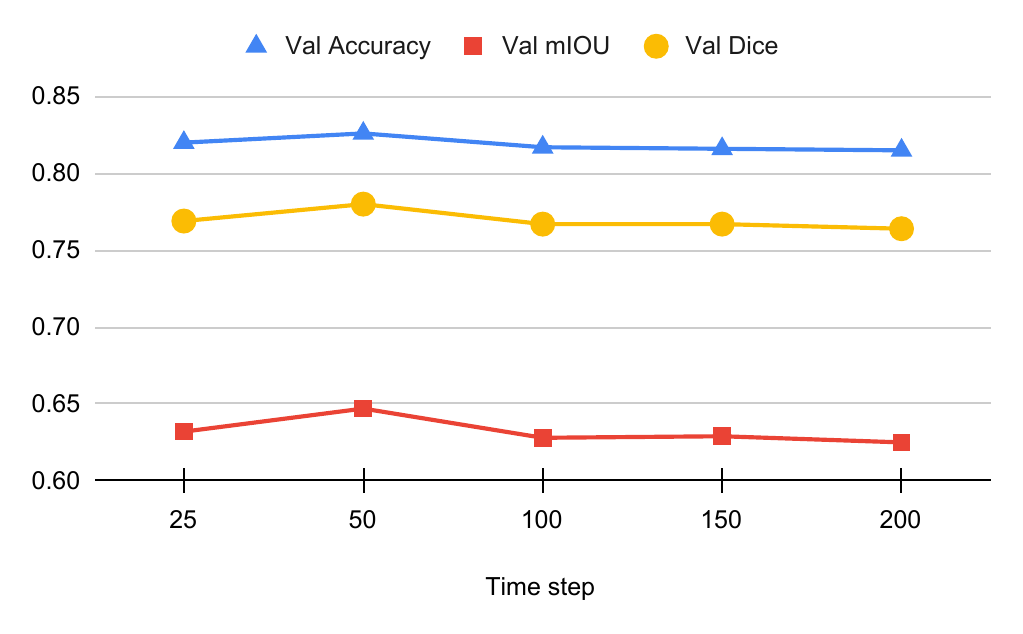}
        \caption{Validation metrics over each timestep}
        \label{fig:timestep}
    \end{minipage}\hfill 
    \begin{minipage}{0.45\textwidth}
        \centering
        \includegraphics[width=\linewidth]{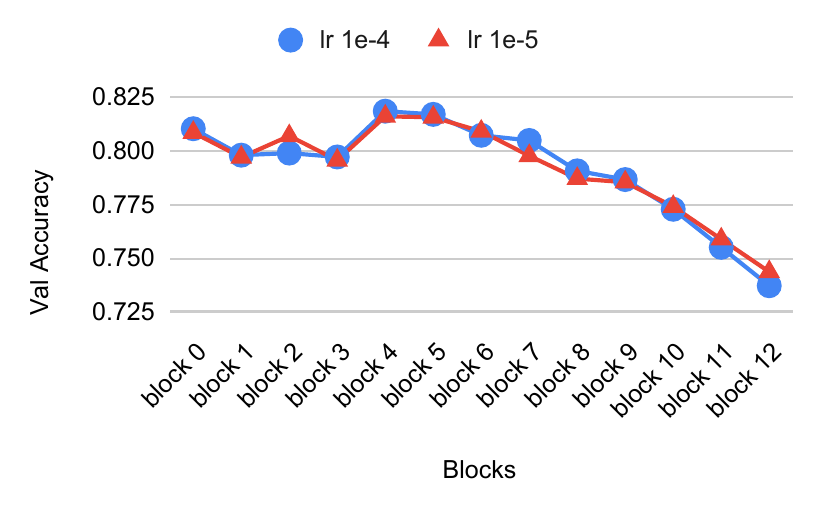}
        \caption{Validation Accuracy of independent blocks}
        \label{fig:blocks}
    \end{minipage}
\end{figure}

To further investigate the impact of feature extraction from different UNet layers within the SSL-guided LDM framework, we conducted ablation studies as shown in Fig. \ref{fig:blocks}. Early-middle layers (e.g., blocks 4–6) consistently achieve higher validation accuracy compared to shallow or deeper layers, indicating that these layers capture optimal structural and semantic features for segmentation tasks. Best accuracy is obtained when all layers are used.

\section{Conclusion}
In this paper, we introduced a novel segmentation approach leveraging a Latent Diffusion Model (LDM) pre-trained on pathology datasets, demonstrating significant improvements over the traditional models pre-trained on ImageNet. Our method outperforms the baselines in BCSS and GlaS datasets, as evidenced by higher Accuracy, Dice, mIOU, and F1 scores. This highlights the effectiveness of domain-specific unsupervised pre-training in capturing intricate tissue structures and enhancing segmentation accuracy.

The qualitative analysis further supports these findings, showcasing our model's ability to maintain structural coherence and accurately delineate complex glandular and tissue boundaries. The proposed architecture advances the state-of-the-art in histopathology image analysis and underscores the importance of tailoring pre-training strategies to specific medical imaging domains. Future work will explore extending this approach to multi-magnification segmentation tasks and optimizing computational efficiency.

\bibliographystyle{unsrt}  
\bibliography{bibliography}

\end{document}